\documentclass[letterpaper]{article}
\usepackage{aaai2026}
\usepackage{times}
\usepackage{helvet}
\usepackage{courier}
\usepackage[hyphens]{url}
\usepackage{graphicx}
\usepackage{natbib}
\usepackage{caption}
\usepackage{amsmath}
\usepackage{amssymb}
\title{Decomposing Theory of Mind: How Emotional Processing Mediates ToM Abilities in LLMs}

\author{
    Ivan Chulo, Ananya Joshi
}
\affiliations{
    Johns Hopkins University\\
    Baltimore, MD 21218 USA\\
}

\begin{document}

\maketitle

\begin{abstract}
Recent work shows activation steering substantially improves language models' Theory of Mind (ToM) \cite{Bortoletto2024}, yet the mechanisms of \textit{what} changes occur internally that leads to different outputs remains unclear. We propose decomposing ToM in LLMs by comparing steered versus baseline LLMs' activations using linear probes trained on 45 cognitive actions. We applied Contrastive Activation Addition (CAA) steering to Gemma-3-4B and evaluated it on 1,000 BigToM forward belief scenarios \cite{gandhi2023understanding}, we find improved performance on belief attribution tasks (32.5\% to 46.7\% accuracy) is mediated by activations processing emotional content : \textit{emotion\_perception} (+2.23), \textit{emotion\_valuing} (+2.20), while suppressing analytical processes: \textit{questioning} (-0.78), \textit{convergent\_thinking} (-1.59). This suggests that successful ToM abilities in LLMs are mediated by emotional understanding, not analytical reasoning.
\end{abstract}

\section{Introduction}

The capacity to attribute mental states to oneself and others represents a critical capability for AI systems engaged in social reasoning, alignment, and human collaboration. Recent benchmarking work by \citet{Bortoletto2024} has demonstrated that activation steering techniques, particularly Contrastive Activation Addition (CAA), can substantially improve language models' ToM performance on belief attribution tasks, often achieving accuracy improvements exceeding 10 \%. This raises a question:\textit{what cognitive processes change} when models successfully engage in perspective-taking?

Traditional evaluations treat ToM as a monolithic capability measured through binary accuracy on belief attribution tasks \cite{gandhi2023understanding}. However, cognitive science views ToM as comprising multiple interacting processes \cite{Gabriel2019}. If steering vectors modulate these components differently, comparing steered versus baseline activation patterns could reveal which processes are \textit{essential} for successful perspective-taking, thus decomposing ToM processes in LLMs into their constituent building blocks.

We introduce a decomposition approach using techniques from mechanistic interpretability in LLMs by combining linear classifer probes trained on 45 cognitive actions with CAA steering vectors. By measuring which cognitive processes systematically increase or decrease when ToM performance improves, we can identify what cognitive actions are correlated with language models' successful perspective-taking.

\textbf{Contributions:} (1) A mechanistic decomposition method using cognitive action probes to analyze steering effects (2) Evaluation on 1,000 BigToM forward belief scenarios showing 14.2\% accuracy improvement (32.5\% to 46.7\%) (3) Analysis demonstrating a systematical increase in emotional/creative processes during perspective-taking and decrease in analytical processes; (4) Findings suggesting that emotional understanding serves as the fundamental building block of perspective-taking in language models.

\begin{figure*}[t]
    \centering
    \includegraphics[width=0.80\textwidth]{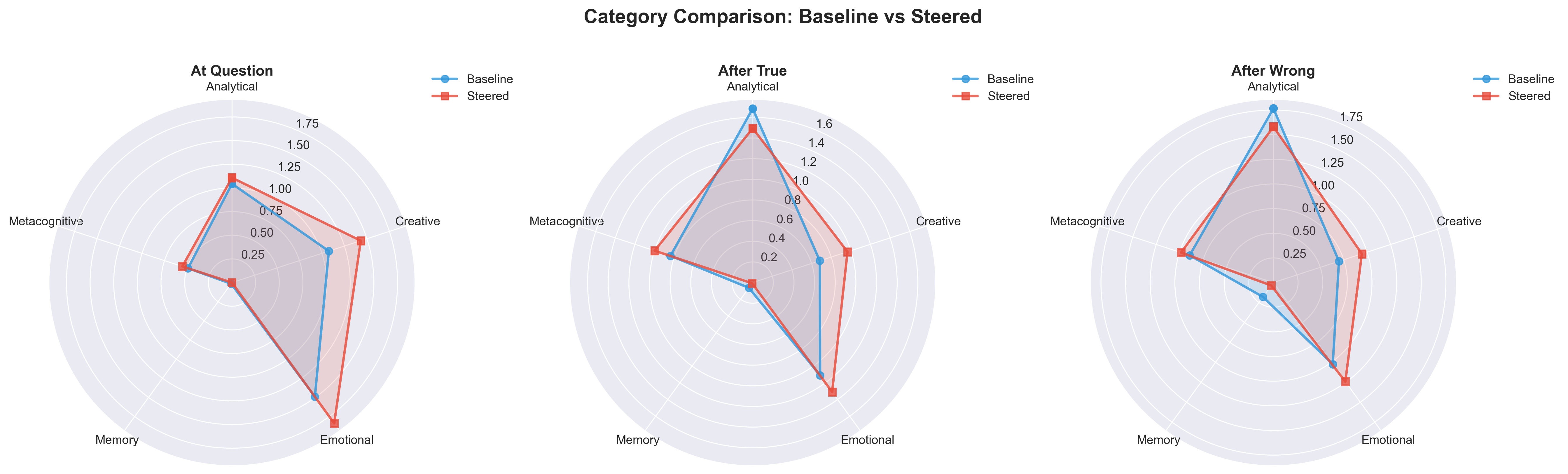}
    \caption{Radar chart comparing baseline versus steered cognitive action activation patterns across categories. The steered condition (red) compared to baseline (blue). These findings support LLMs mirror known cognitive phenomena that emotional understanding is more important than analytical procecsses in perspective-taking }
    \label{fig:radar_comparison_top}
    \end{figure*}

\section{Methodology}
We investigate how steering vectors modulate cognitive processes during belief attribution tasks. 

We developed a multi-stage pipeline integrating probe training, activation steering, and comparative analysis. We began by defining 45 cognitive actions across four categories: \textit{Metacognitive}, \textit{Analytical}, \textit{Creative}, and \textit{Emotional} (see Appendix). For each action, we generated 700 synthetic training samples (31,500 total) of first-person narratives (2-4 sentences) across 20 everyday contexts,   using Gemma-3-4B\footnote{OpenSource link omitted for purposes of review.}. 

Activations were extracted from layers 0-30 of Gemma-3-4B using nnsight, loosely based on the methodology from \cite{chen2024designing}. To ensure consistent probe training, inputs were augmented (at extraction and during inference) with the suffix ``The cognitive action being demonstrated here is'' for consistent final-token extraction. We trained 45 binary linear probes using one-vs-rest classification with AdamW optimization, cosine annealing, and early stopping based on AUC-ROC.

For activation steering, we trained CAA vectors \cite{rimsky2023steering,zou2023representation} on 752 contrastive triplets from BigToM's \cite{gandhi2023understanding} \textit{forward\_belief} scenarios (50/50 false-true split). Each triplet contains story context, belief question, and positive (correct ToM) versus negative (incorrect ToM) completions, capturing representational differences between accurate and inaccurate perspective-taking. Vectors were trained across layers 14-30 using PCA-centered activation differences.

\textbf{Evaluation} We evaluated 1,000 forward belief scenarios from BigToM \cite{gandhi2023understanding} (\textit{forward\_belief\_false}), comparing baseline versus steered conditions. Answers were evaluated by computing p(correct) vs p(incorrect) from model logits. Cognitive action activations were captured at three timepoints: (1) at question, (2) after true answer, (3) after wrong answer. For each action, we computed layer count (layers 10-20 where probe confidence indicated presence) and analyzed baseline-steered differences to identify which cognitive processes characterize successful ToM improvements.

\section{Results}

We first validated our probe methodology, finding that binary probes achieved 0.78 average AUC-ROC and 0.68 F1 across 45 actions. Mid-layer performance (layers 5-24) outperformed early/late layers, informing our layer 10-20 analysis window. We then evaluated CAA steered LLMs on 1,000 forward belief scenarios (false condition), observing accuracy improvements from 32.5\% to 46.7\% (a 14.2\% gain), shifting 217 examples from incorrect to correct predictions.

To understand what drives this improvement, we analyzed cognitive action activation patterns in baseline versus steered conditions. The most striking finding was robust increase of emotional and generative processes. Emotional processes showed strong increases: \textit{emotion\_perception} (mean $\Delta$=+1.73), \textit{emotion\_valuing} ($\Delta$=+0.85), and \textit{emotion\_understanding} ($\Delta$=+0.77). Additionally, \textit{Hypothesis\_generation} ($\Delta$=+1.63) remained strongly elevated across all timepoints, indicating active belief formation and explanation generation.

In contrast, analytical processes decreased: \textit{questioning} ($\Delta$=-1.24), \textit{convergent\_thinking} ($\Delta$=-1.13), and \textit{understanding} ($\Delta$=-0.77), suggesting successful persepective taking in LLMs suppresses deliberate analytical interrogation. Category aggregation reinforces this pattern—creative processes (+0.35, +0.28, +0.24) and emotional processes (+0.35, +0.20, +0.22) consistently increase across timepoints, while analytical processes show decreases (+0.06, -0.19, -0.19).

\section{Discussion}

This pattern of increased emotional and generative processes and decreased analytical processes challenges assumptions that LLM social reasoning relies on deliberate chain-of-thought mechanisms. Rather, successful perspective-taking appears to operate through activating representations responsible for processing emotional contexts.

We make no claims that these findings generalize to human cognition, nor does our methodology validate such comparisons. However, neuroscience research shows that affective and cognitive ToM share neural mechanisms in humans: \citet{Corradi2014} demonstrate that ``patterns in TPJ (Temporoparietal Junction) and MTG (Middle Temporal Gyrus) reflect the same neuronal activity, equally recruited in these two independent conditions.'' During language modeling, networks may learn shared representations linking perspective-taking with emotional context processing, mirroring the compressed structure of human social cognition embedded in linguistic data. Whether this constitutes genuine emulation of cognitive architecture or emergent convergence on functionally equivalent representations remains an open question.

Future work should validate these cognitive decomposition findings with multiple, bigger models and with additional data sources.

\section{Conclusion}

Our work extends \citet{Bortoletto2024} by applying cognitive action probes to activation steered LLMs, we move beyond evaluating \textit{whether} steering improves belief attribution to understanding \textit{why}. Our analysis of 1,000 forward belief scenarios reveals systematic modulation of cognitive processes: steering amplifies generative hypothesis formation and emotional inference while suppressing analytical interrogation. This decomposition approach, combining interpretability tools with targeted interventions, offers a principled methodology for understanding complex AI capabilities. The findings challenge assumptions about deliberative social reasoning in LLMs and open new directions for mechanistic analysis of perspective-taking and other high-level cognitive abilities.

\bibliography{cogni_map_workshop}

\appendix

\section{Appendix}

\subsection{Complete Cognitive Action Taxonomy}

The 45 cognitive actions used for probe training were derived from established cognitive psychology and emotion research frameworks, then organized into five functional categories. Actions were systematically sampled from Bloom's Taxonomy \cite{Anderson2001}, Guilford's Structure of Intellect \cite{Guilford1967}, Metacognitive Processes Framework \cite{Flavell1979}, Krathwohl's Affective Domain \cite{Krathwohl1964}, Gross's Process Model of Emotion Regulation \cite{Gross1998}, and the Mayer-Salovey Emotional Intelligence Model \cite{MayerSalovey1997}.

\textbf{Metacognitive (7 actions):}
\begin{itemize}
    \item \textit{reconsidering}: reconsidering a belief or decision
    \item \textit{updating\_beliefs}: updating mental models or beliefs
    \item \textit{suspending\_judgment}: suspending judgment and staying with uncertainty
    \item \textit{meta\_awareness}: reflecting on one's own thinking process
    \item \textit{metacognitive\_monitoring}: tracking one's own comprehension
    \item \textit{metacognitive\_regulation}: adjusting thinking strategies
    \item \textit{self\_questioning}: interrogating one's own understanding
\end{itemize}

\textbf{Analytical (16 actions):}
\begin{itemize}
    \item \textit{noticing}: noticing a pattern, feeling, or dynamic
    \item \textit{pattern\_recognition}: recognizing recurring patterns across situations
    \item \textit{zooming\_out}: zooming out for broader context
    \item \textit{zooming\_in}: zooming in on specific details
    \item \textit{questioning}: questioning an assumption or belief
    \item \textit{abstracting}: abstracting from specifics to general patterns
    \item \textit{concretizing}: making abstract concepts concrete and specific
    \item \textit{connecting}: connecting disparate ideas or experiences
    \item \textit{distinguishing}: distinguishing between previously conflated concepts
    \item \textit{perspective\_taking}: taking another's perspective or temporal view
    \item \textit{convergent\_thinking}: finding the single best solution
    \item \textit{understanding}: interpreting and explaining meaning
    \item \textit{applying}: using knowledge in new situations
    \item \textit{analyzing}: breaking down into components
    \item \textit{evaluating}: making judgments about value or effectiveness
    \item \textit{cognition\_awareness}: becoming aware and comprehending
\end{itemize}

\textbf{Creative (6 actions):}
\begin{itemize}
    \item \textit{creating}: generating new ideas or solutions
    \item \textit{divergent\_thinking}: generating multiple creative solutions
    \item \textit{hypothesis\_generation}: generating possible explanations
    \item \textit{counterfactual\_reasoning}: engaging in 'what if' thinking
    \item \textit{analogical\_thinking}: drawing analogies between domains
    \item \textit{reframing}: reframing a situation or perspective
\end{itemize}

\textbf{Emotional (15 actions):}
\begin{itemize}
    \item \textit{emotional\_reappraisal}: reinterpreting emotional meaning
    \item \textit{emotion\_receiving}: becoming aware of emotions
    \item \textit{emotion\_responding}: actively engaging with emotions
    \item \textit{emotion\_valuing}: attaching worth to emotional experiences
    \item \textit{emotion\_organizing}: integrating conflicting emotions
    \item \textit{emotion\_characterizing}: aligning emotions with core values
    \item \textit{situation\_selection}: choosing emotional contexts deliberately
    \item \textit{situation\_modification}: changing circumstances to regulate emotion
    \item \textit{attentional\_deployment}: directing attention for emotional regulation
    \item \textit{response\_modulation}: modifying emotional expression
    \item \textit{emotion\_perception}: identifying emotions in self/others
    \item \textit{emotion\_facilitation}: using emotions to enhance thinking
    \item \textit{emotion\_understanding}: comprehending emotional complexity
    \item \textit{emotion\_management}: regulating emotions in self/others
    \item \textit{accepting}: accepting and letting go of control
\end{itemize}

\textbf{Memory (1 action):}
\begin{itemize}
    \item \textit{remembering}: recalling relevant information or experiences
\end{itemize}

\subsection{Synthetic Training Data Generation}

To train cognitive action probes, we generated 31,500 synthetic examples (700 examples per cognitive action) using Gemma-3-4B. Each example consists of a first-person narrative demonstrating a specific cognitive action in an everyday context.

\textbf{Domain Coverage:} Examples were distributed across 20 everyday domains to ensure variety and prevent overfitting to specific contexts: work, school, daily life, cooking, shopping, exercise, reading, writing, planning, learning, organizing, problem-solving, hobbies, personal goals, time management, finances, health, relationships, home projects, and travel.

\textbf{Generation Prompt Template:} For each example, the language model received the following structured prompt:

\begin{verbatim}
Generate a simple, first-person example of
someone [cognitive action].

Action: [action name]
Description: [action description]
Domain: [randomly selected domain]

Requirements:
- Write in first person (I, my, me)
- Keep it simple and realistic
- 2-4 sentences maximum
- Focus on the [action] cognitive action
- Use everyday language

Example only (no explanation):
\end{verbatim}

\textbf{Design Rationale:} First-person narratives capture the internal phenomenology of cognitive processes, providing more direct signal for probe training than third-person descriptions. The 2-4 sentence constraint ensures examples remain focused on a single cognitive action while maintaining sufficient context. Domain randomization prevents the model from learning spurious correlations between cognitive actions and specific contexts (e.g., associating \textit{analyzing} exclusively with academic scenarios).

This approach yielded a balanced dataset of 31,500 training examples with realistic language patterns and diverse contextual grounding, suitable for training binary classifiers to detect cognitive action presence in model activations.

\subsection{CAA Training Data Example}

The CAA steering vectors were trained on contrastive pairs distinguishing correct from incorrect belief attribution. Each training triplet consists of a story with observational information, a question about the protagonist's belief, and two completions representing accurate vs. inaccurate Theory of Mind reasoning.

\textbf{False Belief Example:}

\textit{Story:} ``Noor is working as a barista at a busy coffee shop. Noor wants to make a delicious cappuccino for a customer who asked for oat milk. Noor grabs a milk pitcher and fills it with oat milk. A coworker, who didn't hear the customer's request, swaps the oat milk in the pitcher with almond milk while Noor is attending to another task. Noor does not see her coworker swapping the milk.''

\textit{Question:} ``Does Noor believe the milk pitcher contains oat milk or almond milk?''

\textit{Positive (correct ToM):} ``Noor believes the milk pitcher contains oat milk.''

\textit{Negative (incorrect ToM):} ``Noor believes the milk pitcher contains almond milk.''

\subsection{Evaluation Methodology: Answer Ranking by Probability}

Following the BigToM evaluation protocol, we used answer ranking by probability rather than free-form text generation. Each question was formatted as a multiple-choice task with randomized answer positions:

\textbf{Prompt Format:}
\begin{verbatim}
Story: [story text]

Question: [question text]
Choose one of the following:
a) [answer 1]
b) [answer 2]

Please answer with the letter of your choice (a or b).
Answer:
\end{verbatim}

\textbf{Evaluation Process:}
For each question, we:
\begin{enumerate}
    \item Randomized the positions of true and wrong answers between options a) and b)
    \item Calculated $p(\text{letter}=\text{`a'})$ and $p(\text{letter}=\text{`b'})$ from model logits at the final token position
    \item Selected the answer corresponding to the letter with higher probability
    \item Determined correctness by comparing the selected answer to the ground truth
\end{enumerate}

This probability-based ranking approach eliminates confounds from text generation artifacts and provides a more reliable measure of the model's belief attribution capabilities. Answer position randomization ensures that the model cannot exploit systematic biases in option ordering.

\begin{figure*}[t]
\centering
\includegraphics[width=\textwidth]{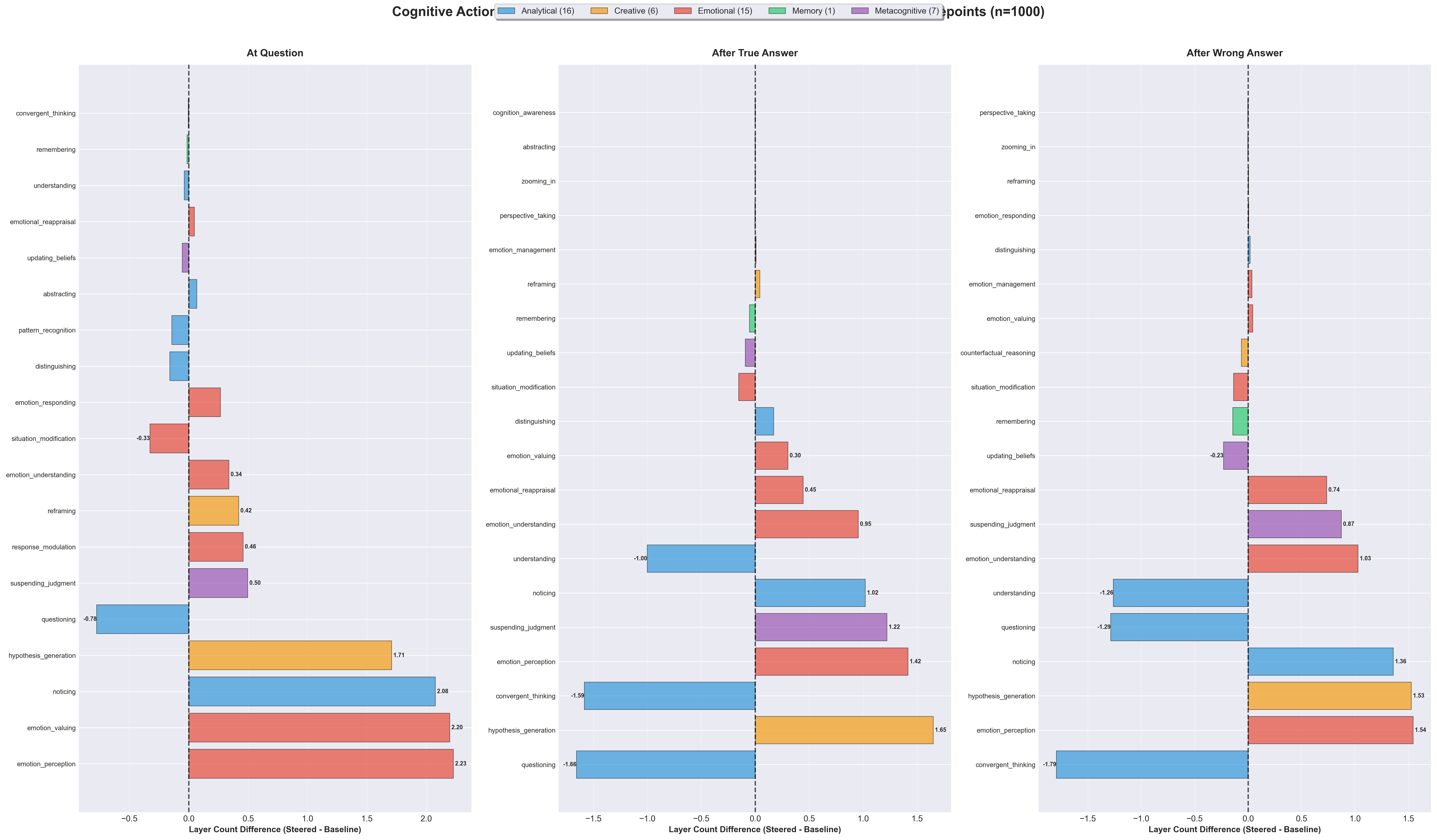}
\caption{Steering effects across all cognitive actions and timepoints (n=1000). Left panels show individual action changes at three timepoints: at question (before answer), after true answer, and after wrong answer. Bars indicate mean layer count difference (steered - baseline) with positive values (right) showing increases and negative values (left) showing decreases. Emotional actions (\textit{emotion\_perception}, \textit{emotion\_valuing}, \textit{noticing}) consistently increase across timepoints, while analytical actions (\textit{questioning}, \textit{convergent\_thinking}, \textit{understanding}) consistently decrease, revealing the emotional foundation of successful ToM.}
\label{fig:complete_comparison}
\end{figure*}

\begin{figure*}[t]
\centering
\includegraphics[width=\textwidth]{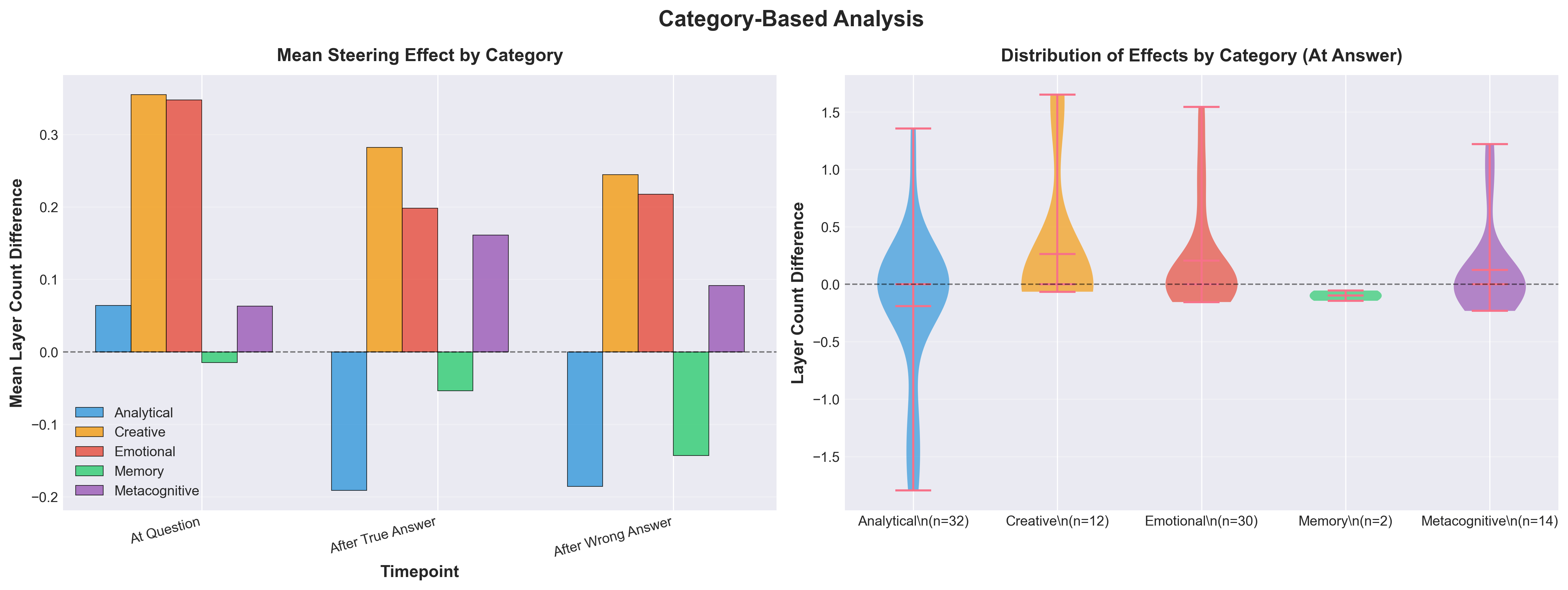}
\caption{Category-level analysis of steering effects. Negative points represent more activation at baseline while positive represent more activation on steered. Left: mean steering effect by cognitive action category. Right: distribution of effects at answer timepoint.}
\label{fig:category_analysis}
\end{figure*}

\begin{figure}[t]
\centering
\includegraphics[width=\textwidth]{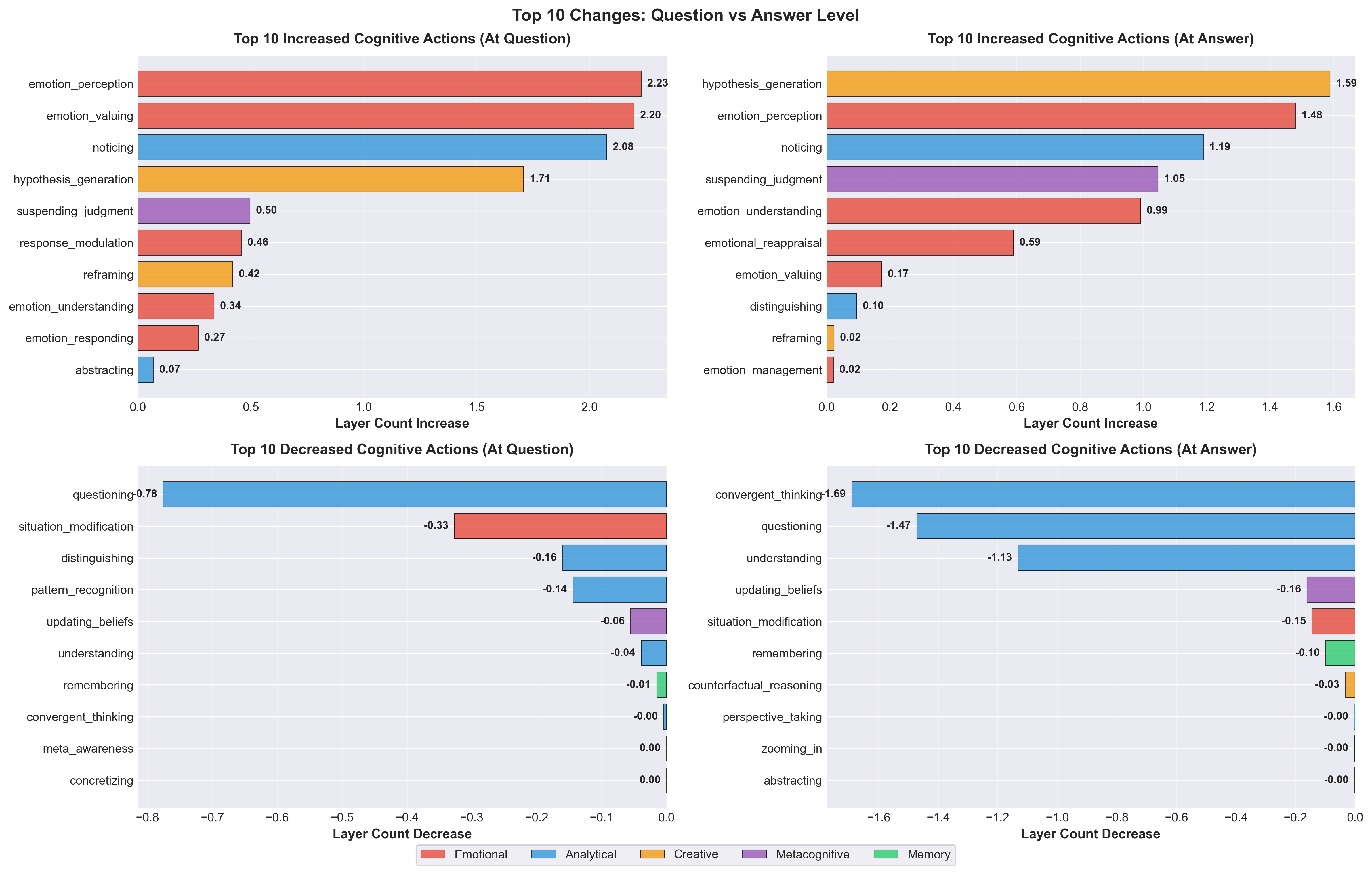}
\caption{Top 10 cognitive actions with largest increases and decreases between baseline and steered conditions at question level (left) and answer level (right). Emotional actions (emotion\_perception, emotion\_valuing, noticing) show the strongest increase, while analytical actions (questioning, convergent\_thinking, understanding) show the strongest decrease, revealing the cognitive processes most affected by successful ToM steering.}
\label{fig:top_changes}
\end{figure}

\begin{figure*}[t]
\centering
\includegraphics[width=\textwidth]{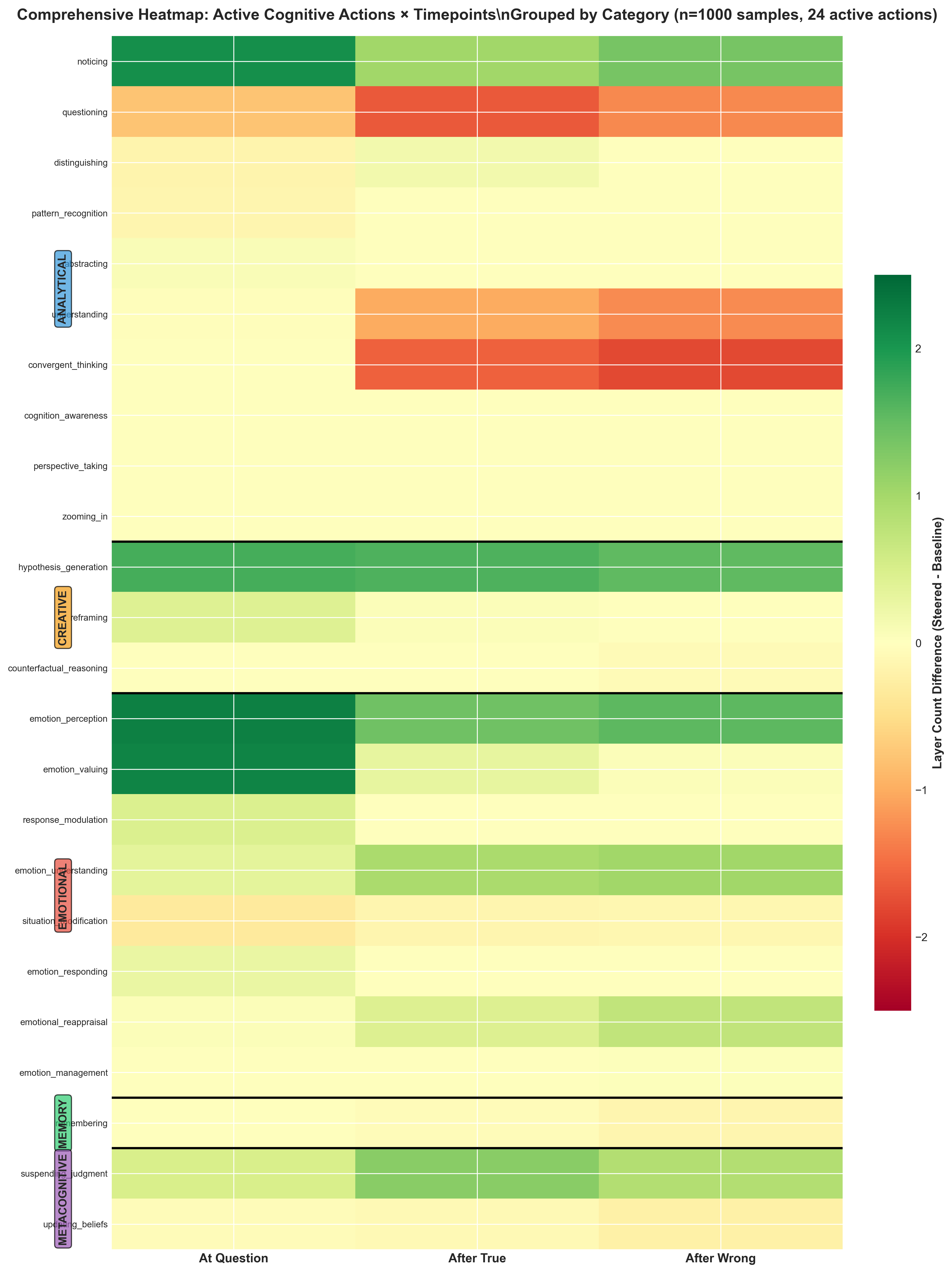}
\caption{Heatmap of cognitive action activation differences (steered - baseline) across timepoints. Each row represents a cognitive action, and columns show the three measurement timepoints: at question, after true answer, and after wrong answer. Green indicates increase and red indicates decreases.}
\label{fig:heatmap}
\end{figure*}

\begin{figure*}[t]
\centering
\includegraphics[width=\textwidth]{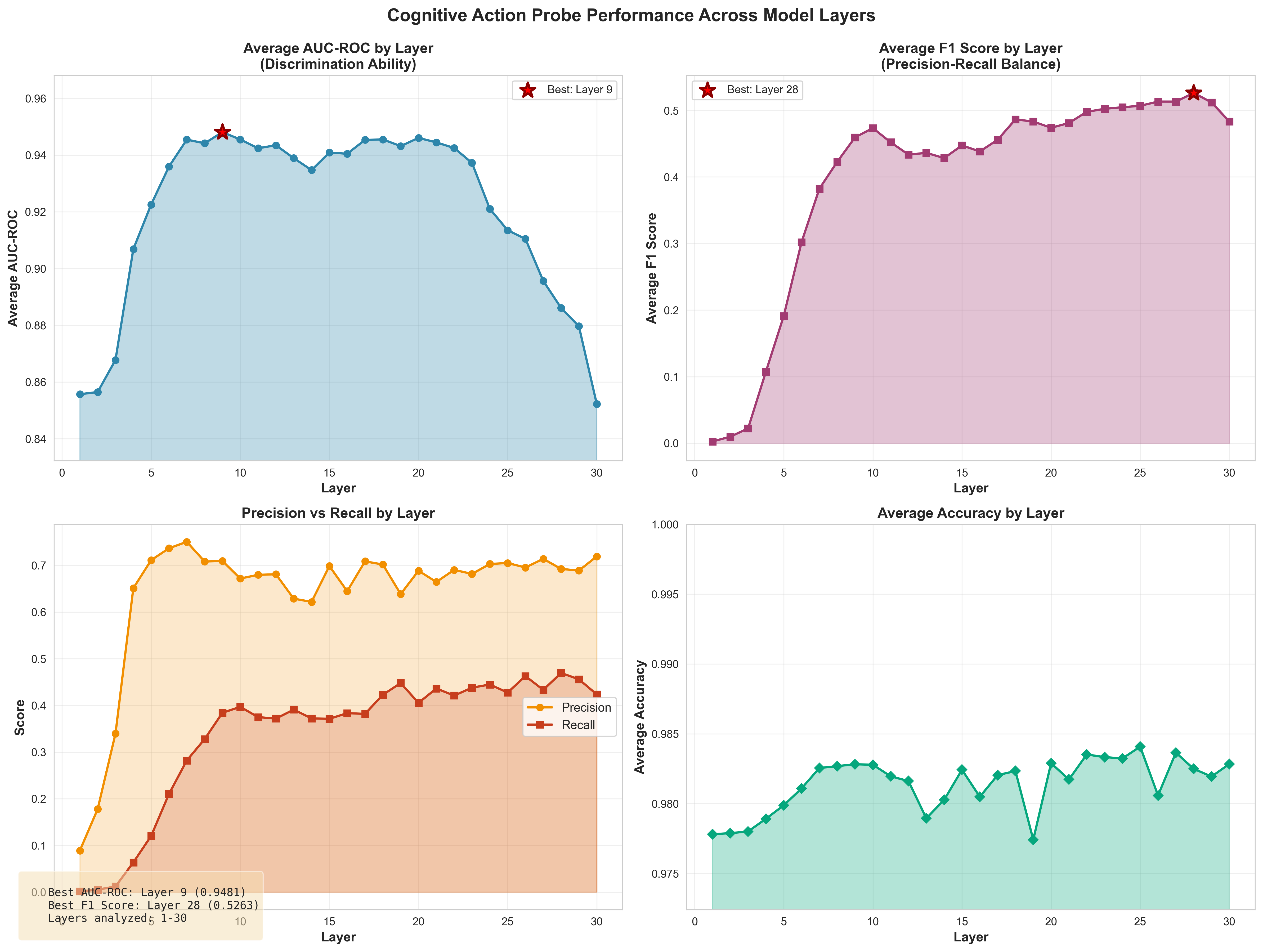}
\caption{Cognitive action probe performance across all 30 layers of Gemma-3-4B. The visualization shows average AUC-ROC scores for each layer, with Layer 9 achieving peak performance (0.948 AUC-ROC). Strong performance is maintained across mid-layers (5-24), while early and late layers show degraded performance. This pattern suggests that early layers focus on surface-level features, mid-layers capture high-level cognitive abstractions, and late layers optimize for next-token prediction, potentially overwriting intermediate representations.}
\label{fig:layer_performance}
\end{figure*}

\begin{figure*}[t]
\centering
\includegraphics[width=\textwidth]{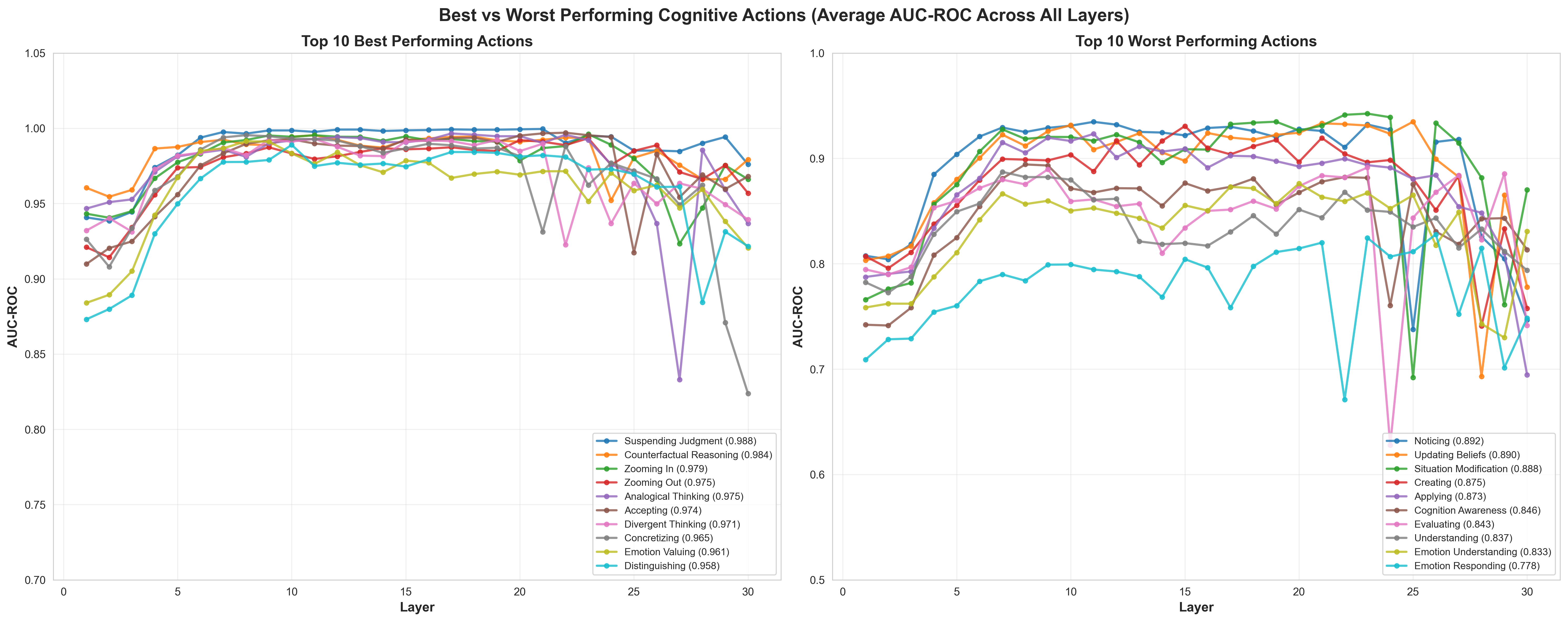}
\caption{Comparison of top 10 and bottom 10 performing cognitive actions ranked by average AUC-ROC across all layers. Best performers like \textit{suspending\_judgment} (0.988) and \textit{counterfactual\_reasoning} (0.984) show consistently high performance and distinct activation patterns across most layers. Worst performers like \textit{emotion\_responding} (0.778) and \textit{understanding} (0.837) exhibit more variability and lower overall discrimination ability, suggesting these concepts may be more distributed or context-dependent in the model's representation space.}
\label{fig:best_worst}
\end{figure*}

\end{document}